\documentclass[10pt,twocolumn,letterpaper]{article}

\usepackage{wacv}
\usepackage{times}
\usepackage{epsfig}
\usepackage{graphicx}
\usepackage{amsmath}
\usepackage{amssymb}
\usepackage{url}
\usepackage{enumitem}

\wacvfinalcopy 

\def\wacvPaperID{****} 

\ifwacvfinal\fi
\begin{document}

\title{Multidomain Document Layout Understanding using Few Shot Object Detection}

\author{Pranaydeep Singh \thanks{Both authors contributed equally}\hspace{0.1cm} \thanks{Work done during an internship at Paralleldots, Inc.} \\
BITS-Pilani, Goa\\
{\tt\small f20140946@goa.bits-pilani.ac.in}
\and
Srikrishna Varadarajan \footnotemark[1] \\
Paralleldots, Inc.\\
{\tt\small srikrishna@paralleldots.com}
\and
Ankit Narayan Singh\\
Paralleldots, Inc.\\
{\tt\small ankit@paralleldots.com}
\and
Muktabh Mayank Srivastava  \\
Paralleldots, Inc.\\
{\tt\small muktabh@paralleldots.com}
}

\maketitle
\ifwacvfinal\thispagestyle{empty}\fi

\begin{abstract}
We try to address the problem of document layout understanding using a simple algorithm which generalizes across multiple domains while training on just few examples per domain. We approach this problem via supervised object detection method and propose a methodology to overcome the requirement of large datasets. We use the concept of transfer learning by pre-training our object detector on a simple artificial (source) dataset and fine-tuning it on a tiny domain specific (target) dataset. We show that this methodology works for multiple domains with training samples as less as 10 documents. We demonstrate the effect of each component of the methodology in the end result and show the superiority of this methodology over simple object detectors.
\end{abstract}

\section{Introduction}

The understanding of document layout in terms of finding logical components such as title, paragraphs etc. is a preliminary step towards retrieving information from images of documents. The amount of variability in real-world data coming from multiple domains e.g., documents, invoices etc. makes it a challenging computer vision problem that has intrigued researchers for decades.

The most basic version of the layout understanding task is to separate text from background and images, but the task has evolved to not only segregating these basic structures but also derived structures like paragraphs, lists and tables. Various image processing methodologies \cite{docstrum} \cite{Voronoi++} \cite{Namboodiri2007} have approached the problem of understanding general documents as well as digitizing historical documents. With the onset of deep learning and data driven approaches, the problem was approached as a pixel-wise segmentation task \cite{docseg_adobe}, where each pixel is assigned a class based on its surrounding pixels. In this paper, we explore a new tangent, where the problem is approached as a few-shot object detection problem to identify relevant areas in a document. The motivation is to understand document structure with as less as 10 tagged examples since digitization tasks generally don’t have an abundance of tagged data at hand. However, understanding documents is a complicated task and a dataset consisting of just 10 examples is not enough to train an object detector especially (as they're fully supervised networks requiring large amounts of training data) to understand various structures, like tables or lists. 

Hence, we use a transfer learning based approach where we give the network a general understanding of what basic features and structures are contained in a document and then proceed to train on a few-shot task for understanding of specific document types like invoices, resumes, academic papers, journals etc. A few-shot task is described widely as training the model using just a handful of tagged examples. 

 The initial network which is to be later used for fine-tuning needs to have a wide understanding of document structures and substructures and needs to be trained extensively for it to yield good results when fine-tuned with very less samples. There was no relevant dataset which accommodated these needs and hence, we artificially generated a simple dataset using HTML. We refer to this dataset as Source Dataset. We then proceed to train the described model on this dataset. This trained model now serves as the backbone of all future models we fine-tuned. Using as little as 10, and up to 50 images, we demonstrate that the obtained model learns to understand document structures. We also show that the methodology can be extended to any number of domains with few examples from each. In this paper, we demonstrate the methodology and its application to Invoices and Resume images. We call these domains as Target Domains and the datasets as Target Datasets.

Our contributions consists of the following points
\begin{itemize} [noitemsep]
    \item Applying state of the art object detection techniques for Document Layout Understanding
    \item Introducing a generalized algorithm which can perform Layout Understanding in multiple domains using just few tagged images (eg: 10).
\end{itemize}

\begin{figure*}
\begin{center}
\includegraphics[width=0.4\textwidth,height=0.3\textheight]{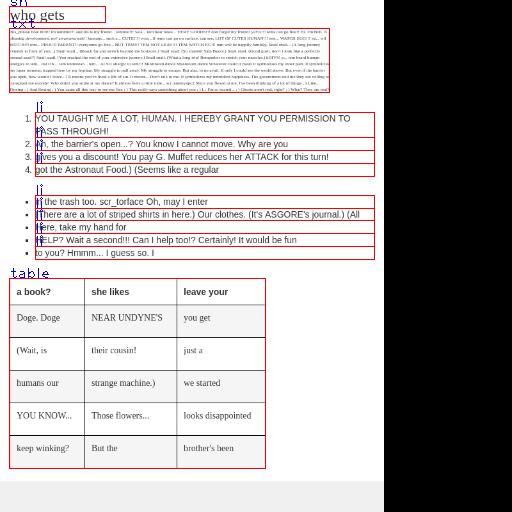}
\hspace{0.1\textwidth}
\includegraphics[width=0.4\textwidth,height=0.2\textheight]{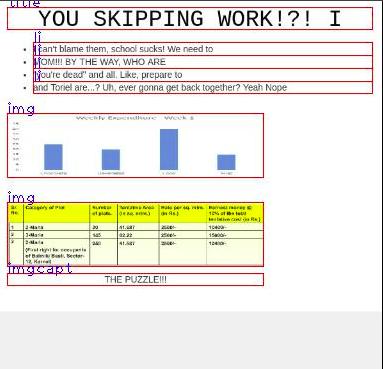}
\end{center}
   \caption{Sample images from the Artificial Dataset}
\label{fig:artificial}
\end{figure*}

\section{Related Work}
There are two sub-parts to the Document Layout Analysis problem
\begin{itemize}
    \item Geometric Layout Analysis
    \item Logical Layout Analysis
\end{itemize}

Geometric Layout Analysis (GLA) is centred around understanding the basic geometric layout of a document, such as skew, page decomposition, text detection etc. Logical Layout Analysis (LLA) focuses on understanding the implied semantic labels in a document, like captions, subheading, table headings etc. GLA has been addressed mainly by image processing methods like Hough Transforms and Binarization. While the GLA problem is as old as Image Processing itself, LLA is a more recent problem and the one which we attempt to solve. Approaches employed in LLA mainly follow the bottom-up approach. Bottom-up approaches work by finding the smallest entities like words or characters and attempt to aggregate them using a distance metric and an aggregation algorithm like K-Nearest Neighbors or K-D Trees. These approaches \cite{docstrum} \cite{Voronoi++} \cite{Namboodiri2007} have the advantage of being mostly unsupervised but involve tuning a lot of heuristics. They are also not scalable to document layouts which are different from those the algorithm is tuned on. Comparisons of such approaches are also covered by \cite{6comparison} \cite{compami}. The most popular and widely used of these approaches is the Docstrum \cite{docstrum} algorithm. It uses KNN to aggregate the minute structures into lines and then employs heuristics like, perpendicular distance and angle between lines to combine them into text blocks. While deep learning approaches to LLA also exist, these approaches \cite{docseg_adobe} \cite{cnnevaldoc} require vast amounts of training data and only learn a fixed set of labels and are thus not useful for few-shot tasks with a wide variety of different labels. We explore an object detection based approach to LLA, which can be fine-tuned on as less as 10 images to understand semantic labels like address, total bill amount, skills, education etc.

Few shot object detection is a task where the tagged training set is very small (say 1-50 images total). Previous work has been explored on the PASCAL VOC/COCO/ImageNet dataset. \cite{lstd} introduce a Low-shot Object Detector (LSTD) model which is pretrained on a huge Source Dataset and fine-tuned on a small (low-shot) target dataset. The LSTD model is based on Single Shot Detector (SSD) \cite{ssd} and Faster-RCNN (FRCNN) \cite{frcnn}. Broadly, they use the SSD network to detect foreground segments and a classifier which takes ROIPooled features from the SSD feature maps to classify the detected regions. There are two regularizations introduced by \cite{lstd}, Background Regularization (BGR) and Tk-Regularization (Tk-R) which helps them in learning from just few examples in the target dataset. The Source dataset in our case is more basic while \cite{lstd} assume the Source dataset to be very huge and comprehensive.

\begin{figure}
\begin{center}
\includegraphics[width=0.5\textwidth]{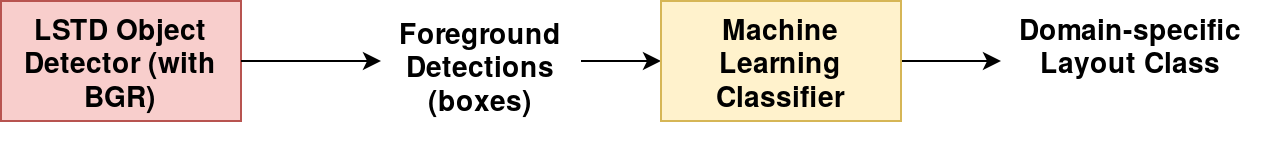}
\end{center}
   \caption{Overview of the proposed method}
\label{fig:method}
\end{figure}

\begin{figure*}
\includegraphics[width=0.5\textwidth,height=0.5\textheight]{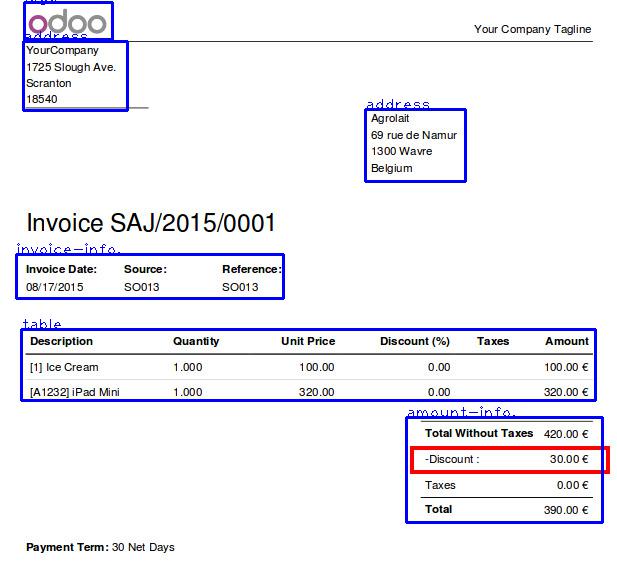}
\includegraphics[width=0.5\textwidth,height=0.5\textheight]{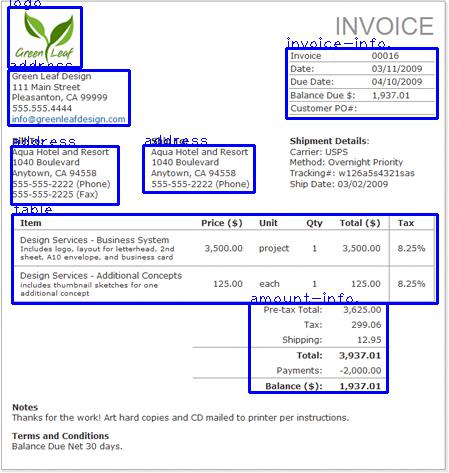}
   \caption{Sample tagged images from the Invoice Dataset}
\label{fig:invoices}
\end{figure*}

\section{Architecture} 
Our architecture is a two-step object detector. The first step is the detector (inspired from LSTD) which detects the foreground regions and the second step is the ML classifier which predicts the domain-specific layout class.

For the first step, we leverage a better feature extractor for the object detector. We use the Feature Pyramid Networks \cite{fpn} as our feature extractor. This (FPN based SSD) achieves state-of-the-art performance for a single model on PASCAL VOC dataset (object detection) as shown here \footnote{\url{https://github.com/kuangliu/torchcv}}.

On the Target dataset, many of the target classes cannot be distinguished by visual features alone. Hence we resorted to using a separate classifier (as opposed to the FRCNN based LSTD classifier) for the detected boxes. This involves taking text based features.  Hence, while fine-tuning, a better alternative to this classifier is used in our system. The learning of target domain is made easier and faster by making use of the background regularization constraint.

\section{Methodology}
The task can be described as few shot document layout understanding. Our methodology consists of the following parts
\begin{enumerate}[noitemsep]
    \item Creating the artificial (Source) dataset.
    \item Pretraining the model on the Source dataset.
    \item Finetuning the model on the domain-specific (Target) dataset.
    \item Training the ML classifier on the Target dataset (is combined with Step 3)
\end{enumerate}

\subsection{Dataset Generation} 

Our artifical dataset contains 160,000 images spanning multiple scales and sizes, accommodating for asymmetrically placed structures and elements. The dataset contained 8 basic layout classes \begin{enumerate}[noitemsep]
    \item Title
    \item Heading
    \item Sub-Heading
    \item Text Block
    \item List
    \item Table
    \item Image Content
    \item Image/Table Caption
\end{enumerate}
The textual content in the dataset was taken from a text dump consisting of a variety of online sources. The images were taken from a small dataset collected from Google Images. Apart from random images, the image dataset contained specific images collected using relevant keywords like graphs, tables, charts etc. Few examples from the artificial dataset along with its taggings are shown in Figure \ref{fig:artificial}

\begin{figure}
\includegraphics[width=0.5\textwidth]{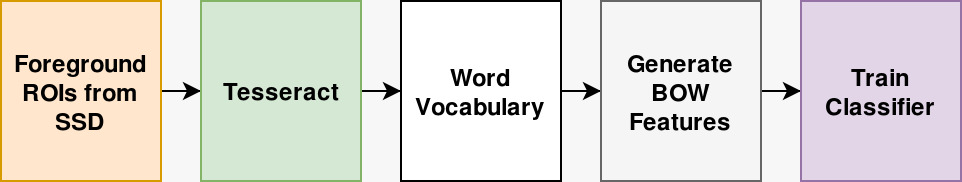}
  \caption{Overview of the ML Classifier}
\label{fig:mlcls}
\end{figure}

\subsection{Training}

We train the LSTD model as it is on the Source Dataset. Once our model is trained on the Source Dataset, we move to fine-tune the model on the Target Datasets. Here we apply BGR. As mentioned earlier, we found that the performance of the inbuilt classifier in LSTD was not performing to our satisfaction, hence we decided to pass the foreground detections from the network through a seperate classifier. 

\textbf{Target Classification:} To tackle the domain specific layout classes, we employed few ways to extract the best features so that we can train a classifier. We extracted the text from the detected box and used bag-of-words approach for getting the textual features. We also used other features related to the spatial configuration of the detected box. We use these features to train a machine learning algorithm to classify the detected bounding box to one of the classes. This is described in Figure \ref{fig:mlcls}

\subsection{Implementation Details}
For creating the artificial dataset, we generated HTML files which correspond to web documents and exported them into images using a webdriver. For the layout detection step, we implemented the LSTD network in PyTorch library. We use the FPNSSD from torchcv library \footnote{\url{https://github.com/kuangliu/torchcv}}. For all experiments, we use SGD optimizer with learning rate of 0.0001 and momentum 0.9. We use L2 penalty of 0.0005.

For the layout classification step, to extract text from a detected box we use the open-source LSTM-based Tesseract 4.0\footnote{\url{https://github.com/tesseract-ocr/tesseract}}. We get our classifier using the tpot toolkit \cite{tpot}, which uses genetic programming to optimize machine learning pipelines.

While reporting the results, we take the IoU threshold for the object detection accuracy metrics as 0.5



\section{Invoice Dataset}

\subsection{Dataset description}
We collected 170 invoices which includes variations in structure, domain and template. We refer to this as the Invoice Dataset. We manually tag this dataset into layouts of 5 main categories - \begin{enumerate}[noitemsep]
    \item Logo 
    \item Address
    \item Bill/Invoice Information
    \item Tables 
    \item (Total) Amount Information
\end{enumerate}

Few example images from the dataset are shown in Fig \ref{fig:invoices}. We use a fixed set of 100 images as our test set. We train our model on different (incremental) number of training images (k) and report the results correspondingly.

\begin{table}[]
\begin{center}
\begin{tabular}{llll}
\hline
\begin{tabular}[c]{@{}l@{}}No of training\\ images (k)\end{tabular} & \begin{tabular}[c]{@{}l@{}}Mean\\ Precision\end{tabular} & \begin{tabular}[c]{@{}l@{}}Mean\\ Recall\end{tabular} & \begin{tabular}[c]{@{}l@{}}Mean\\ F1 Score\end{tabular} \\ \hline
10                                                                   & 0.4721    & 0.5188 & 0.4943   \\ 
20                                                                   & 0.4962    & 0.5444 & 0.5192   \\ 
30                                                                   & 0.5012    & 0.5791 & 0.5373   \\ 
40                                                                   & 0.5244    & 0.601  & 0.5601   \\ 
50                                                                   & 0.5316    & 0.6101 & 0.5682   \\ 
60                                                                   & 0.5599    & 0.6214 & 0.589    \\ 
70                                                                   & 0.56      & 0.6354 & 0.5953   \\ \hline
\end{tabular}
\end{center}
    \caption{LSTD End to End accuracy on \textbf{Invoice Dataset}}
    \label{table:invoice-e2e}
\end{table}

\begin{table}[]
\begin{center}
    \begin{tabular}{llll}
    \hline
    \begin{tabular}[c]{@{}l@{}}No of training\\ images (k)\end{tabular} & Precision & Recall & F1 Score \\ \hline
    0                             & 0.144     & 0.4214 & 0.2147   \\
    10                            & 0.5992    & 0.6212 & 0.61     \\
    20                            & 0.611     & 0.7062 & 0.655    \\
    30                            & 0.6203    & 0.7755 & 0.6893   \\
    40                            & 0.6767    & 0.7901 & 0.729    \\
    50                            & 0.6742    & 0.7992 & 0.7314   \\
    60                            & 0.7017    & 0.8001 & 0.7484   \\
    70                            & 0.7292    & 0.8132 & \textbf{0.7689}   \\ \hline
    \end{tabular}
\end{center}
    \caption{LSTD foreground detection accuracy on \textbf{Invoice Dataset}}
    \label{table:invoice-fg}
\end{table}

\begin{table}[]
\begin{center}
\begin{tabular}{llll}
    \hline
    \begin{tabular}[c]{@{}l@{}}No of training \\ images (k)\end{tabular} & Precision & Recall & F1 score \\ \hline
    10                                                                   & 0.1078    & 0.1991 & 0.1399   \\ 
    20                                                                   & 0.1377    & 0.235  & 0.1736   \\
    30                                                                   & 0.1744    & 0.2768 & 0.214    \\
    40                                                                   & 0.1957    & 0.2998 & 0.2368   \\
    50                                                                   & 0.3018    & 0.3036 & 0.3027   \\
    60                                                                   & 0.3738    & 0.315  & 0.3419   \\
    70                                                                   & 0.3888    & 0.3445 & \textbf{0.3653}   \\ \hline
    \end{tabular}
\end{center}
    \caption{LSTD \textbf{without Source pretraining} on \textbf{Invoice Dataset} - foreground detection accuracy}
    \label{table:invoice-scratch}
\end{table}

\begin{table}[]
\begin{center}
    \begin{tabular}{llll}
    \hline
    \begin{tabular}[c]{@{}l@{}}No of training \\ images (k)\end{tabular} & Precision & Recall & F1 score \\ \hline
    70                                                                   & 0.7718    & 0.8135 & 0.7921   \\
    \end{tabular}
\end{center}
    \caption{ML Classifier accuracy on \textbf{Invoice Dataset}}
    \label{table:invoice-ml}
\end{table}


\begin{figure}
\includegraphics[width=0.5\textwidth,height=0.5\textheight]{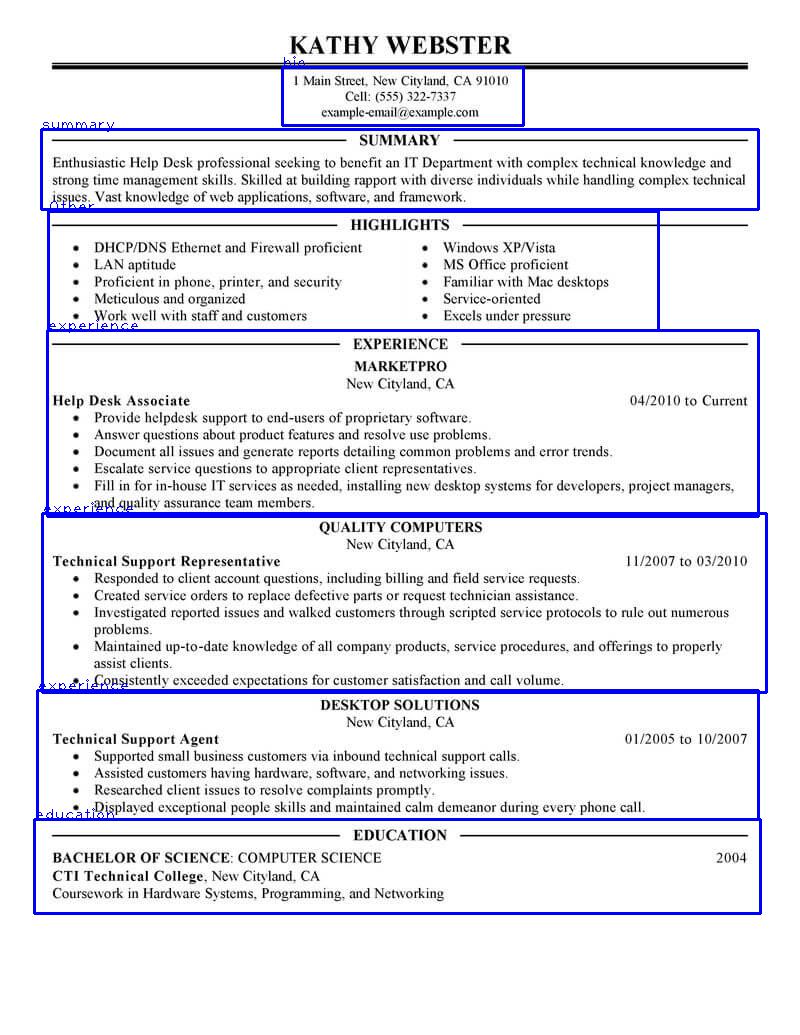}
   \caption{Sample tagged images from the Resume Dataset}
\label{fig:resumes}
\end{figure}

\section{Resume Dataset}
\subsection{Dataset description}
The resume dataset is a set of 100 images collected from various sources containing resumes from different domains and layouts. As with the invoice dataset, this was manually tagged into 6 main categories: 
\begin{enumerate}[noitemsep]
    \item Education 
    \item Experience 
    \item Bio 
    \item Skills 
    \item Summary
    \item Other
\end{enumerate}

Example Images are shown in Fig \ref{fig:resumes}. A fixed set of 50 images is used as the test set and training is done on an incremental number of training images ranging from 10 to 50. 


\begin{table}[]
\begin{center}
\begin{tabular}{llll}
\hline
\begin{tabular}[c]{@{}l@{}}No of training\\ images (k)\end{tabular} & \begin{tabular}[c]{@{}l@{}}Mean\\ Precision\end{tabular} & \begin{tabular}[c]{@{}l@{}}Mean\\ Recall\end{tabular} & \begin{tabular}[c]{@{}l@{}}Mean\\ F1 Score\end{tabular} \\ \hline
10                                                                   & 0.6144    & 0.5888 & 0.6013   \\ 
20                                                                   & 0.6398    & 0.6011 & 0.6198   \\ 
30                                                                   & 0.6587    & 0.6218 & 0.6397   \\ 
40                                                                   & 0.6712    & 0.6325 & 0.6513   \\ 
50                                                                   & 0.6946    & 0.634  & 0.6629   \\ \hline
\end{tabular}
\end{center}
    \caption{LSTD End to End accuracy on \textbf{Resume Dataset}}
    \label{table:resume-e2e}
\end{table}

\begin{table}[]
\begin{center}
    \begin{tabular}{llll}
    \hline
    \begin{tabular}[c]{@{}l@{}}No of training\\ images (k)\end{tabular} & Precision & Recall & F1 Score \\ \hline
    0                        & 0.035     & 0.4311 & 0.06     \\
    10                       & 0.8228    & 0.821  & 0.8219   \\
    20                       & 0.8542    & 0.8224 & 0.838    \\
    30                       & 0.8655    & 0.8291 & 0.8469   \\
    40                       & 0.9123    & 0.8363 & 0.8726   \\
    50                       & 0.8977    & 0.8343 & \textbf{0.8659}   \\ \hline
    \end{tabular}
\end{center}
    \caption{LSTD foreground detection accuracy on \textbf{Resume Dataset}}
    \label{table:resume-fg}
\end{table}

\begin{table}[]
\begin{center}
    \begin{tabular}{llll}
    \hline
    \begin{tabular}[c]{@{}l@{}}No of training \\ images (k)\end{tabular} & Precision & Recall  & F1 score \\ \hline
    10                                                                   & 0.3797    & 0.3571  & 0.368    \\
    20                                                                   & 0.3859    & 0.3928  & 0.3893   \\
    30                                                                   & 0.5238    & 0.5238  & 0.5238   \\
    40                                                                   & 0.5178    & 0.7532  & 0.6137   \\
    50                                                                   & 0.60946   & 0.61309 & \textbf{0.61037}  \\ \hline
    \end{tabular}
\end{center}
    \caption{LSTD \textbf{without Source pretraining} on \textbf{Resume Dataset} - foreground detection accuracy}
    \label{table:resume-scratch}
\end{table}

\begin{table}[]
\begin{center}
    \begin{tabular}{llll}
    \hline
    \begin{tabular}[c]{@{}l@{}}No of training \\ images (k)\end{tabular} & Precision & Recall & F1 score \\ \hline
    50                                                                   & 0.804     & 0.8946 & 0.8469   \\
    \end{tabular}
\end{center}
    \caption{ML Classifier accuracy on \textbf{Resume Dataset}}
    \label{table:resume-ml}
\end{table}


\section{Results}

\begin{table}[]
\begin{center}
    \begin{tabular}{llll}
    \hline
    Dataset & Precision & Recall & F1 score \\ \hline
    Invoice & 0.0547    & 0.1935 & \textbf{0.0853}   \\
    Resume  & 0.2415    & 0.2559 & \textbf{0.2485} 
    \end{tabular}
\end{center}
    \caption{Table: \textbf{Baseline} (Docstrum) accuracy}
    \label{table:baseline}
\end{table}

\begin{figure*}
\includegraphics[width=0.5\textwidth,height=0.5\textheight]{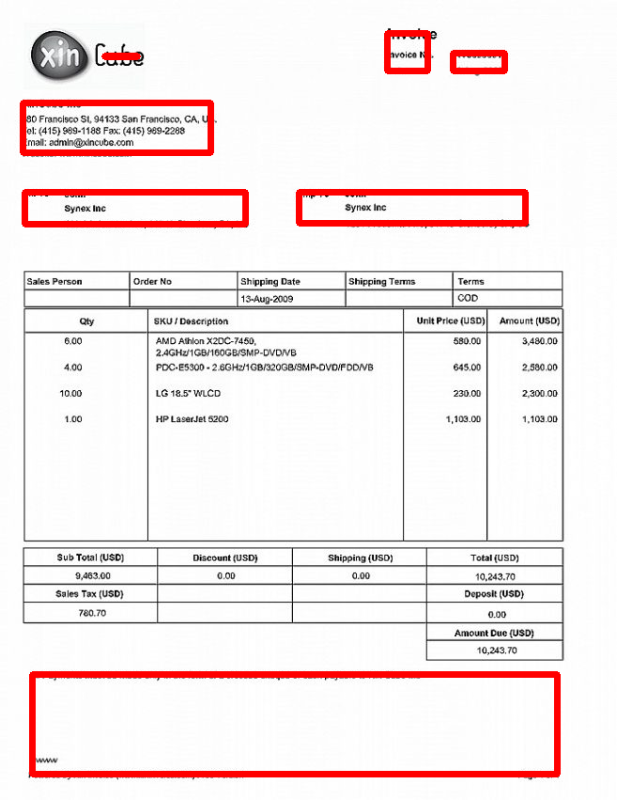}
\includegraphics[width=0.5\textwidth,height=0.5\textheight]{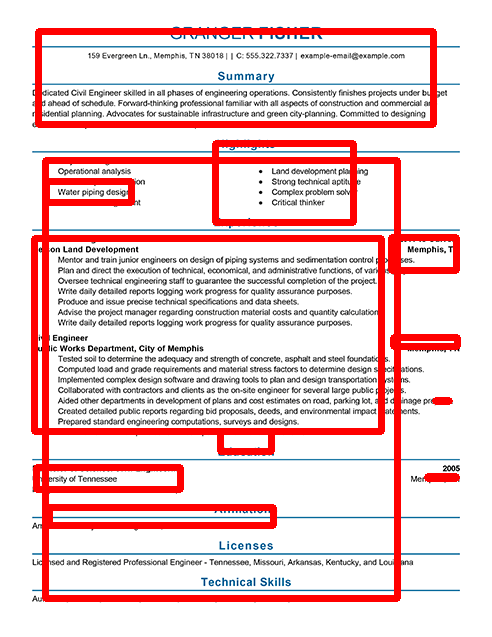}
   \caption{Sample predictions of the baseline method on both Datasets}
\label{fig:baseline}
\end{figure*}

\textbf{Baselines}: The Docstrum algorithm \cite{docstrum} serves as our baseline. The algorithm converts images to grayscale and binarizes them. It further finds the connected components and their centroids. It then looks for the K-nearest neighbours (K=5) of each component. Vectors are plotted from each centroid to its neighbours and these angles help in skew correction. The nearest-neighbor distance histogram has several peaks and these peaks typically represent between-character spacing, between-word spacing and between-line spacing. These values are then used to construct lines, words and text blocks with some predetermined tolerance for each spacing value. 

We use Docstrum to construct blocks and then determine the accuracy using the manually annotated ground truth results on both the target datasets ie. Invoices and Resumes. Sample outputs for the same are shown in Figure \ref{fig:baseline}

Table 1-4 shows the various results on the Invoice Dataset, while Table 5-8 shows the results on the Resume Dataset.

Table \ref{table:invoice-fg}, \ref{table:resume-fg} shows the accuracy of just the foreground detections (LSTD detections) while Table \ref{table:invoice-ml}, \ref{table:resume-ml} shows the accuracy of just the ML Classifier on the foreground ROIs. Table \ref{table:invoice-e2e},\ref{table:resume-e2e} shows the \textbf{end to end accuracy} of both the foreground detection and ML Classifier combined. It is clearly evident from the results that the method works great even for 10 training examples.

Importance of our \textbf{Source pretraining} is shown by the results given in Table \ref{table:invoice-scratch}, \ref{table:resume-scratch}. These are the results from the models which are \textbf{trained from scratch} instead of finetuning the pretrained model on Source dataset. One can notice an accuracy improvement of at least \textbf{40\%} on F1 scores of Target Domain Layout Detection task. 

The above results also demonstrate the superiority of the methodology over simple object detectors.

\section{Discussion}

Interesting observation with regards to Zero Shot Transfer Learning is that the performance mainly depends on how varied one's Source Dataset is. Upon qualitative analysis, we found that the Resume Dataset is more varied than our artifical dataset when compared to Invoice Dataset and our artificial dataset. This explains why even with 0 samples, directly applying Source trained model on the test images of Target Datasets work reasonably in the case of Invoice Dataset. Further improvements can be made in the Source Dataset generation to make it more generalized and varied, will help narrow this gap.

There are two regularizations introduced by \cite{lstd}, Background Regularization (BGR) and Tk-Regularization (Tk-R). We use BGR to make the learning of Target domain easier and faster. This is achieved by making the learning of background part in the Target domain easier through this constraint. Tk-R tries to bridge the gap between predictions of the classifier on Source and Target domain. Tk-R was not useful as the default (FRCNN based) classifier does not perform well due to the reasons mentioned earlier.


\begin{figure*}
\includegraphics[width=0.5\textwidth,height=0.5\textheight]{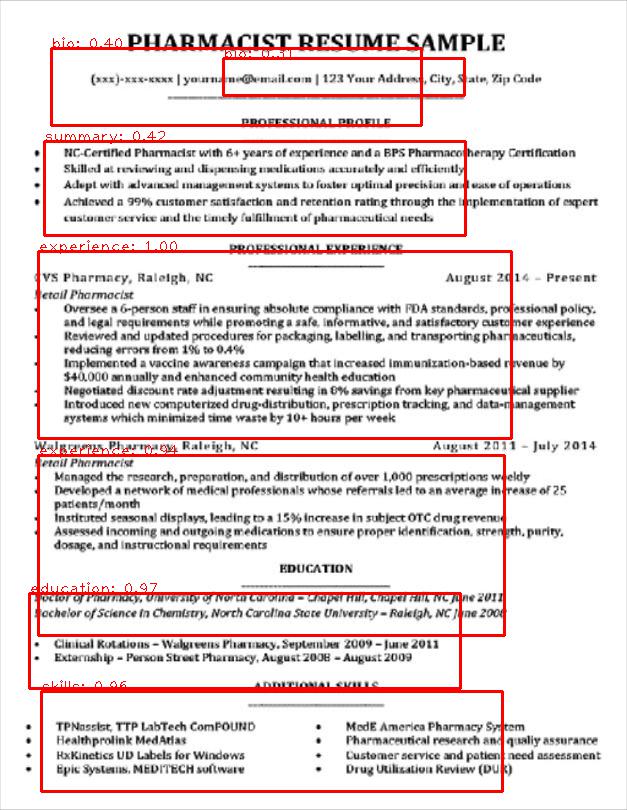}
\includegraphics[width=0.5\textwidth,height=0.5\textheight]{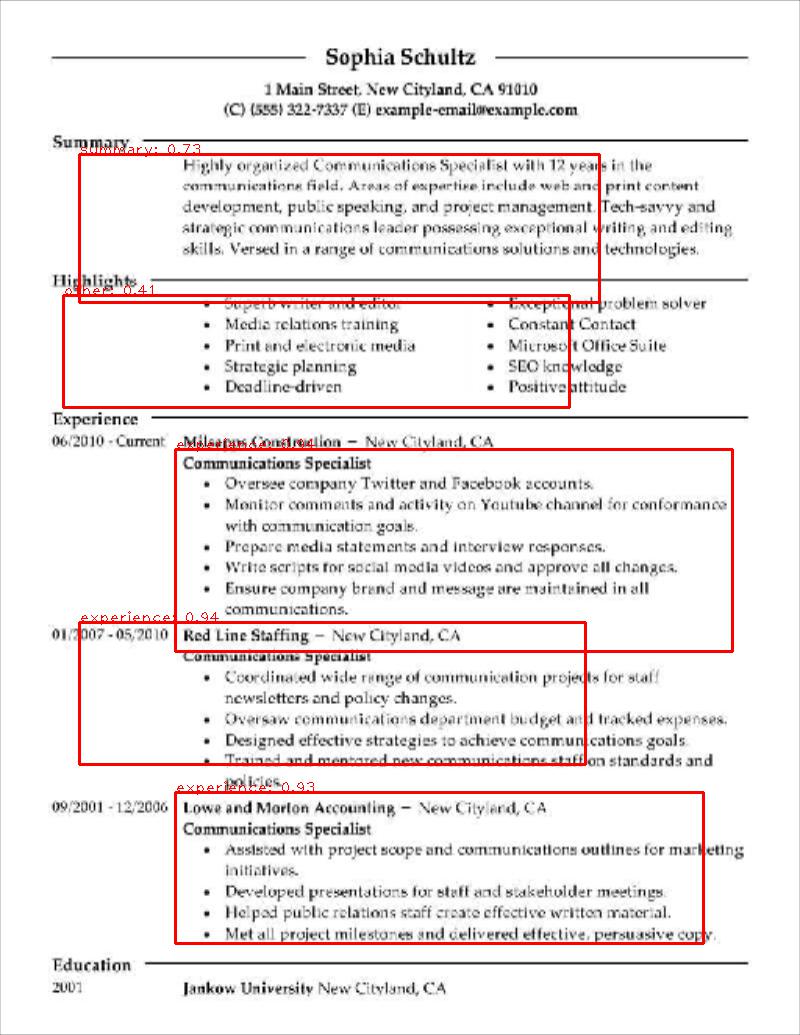}
   \caption{Sample predictions from our system on the test images of Resume Dataset}
\label{fig:resumes_pred}
\end{figure*}

\begin{figure*}
\includegraphics[width=0.5\textwidth,height=0.5\textheight]{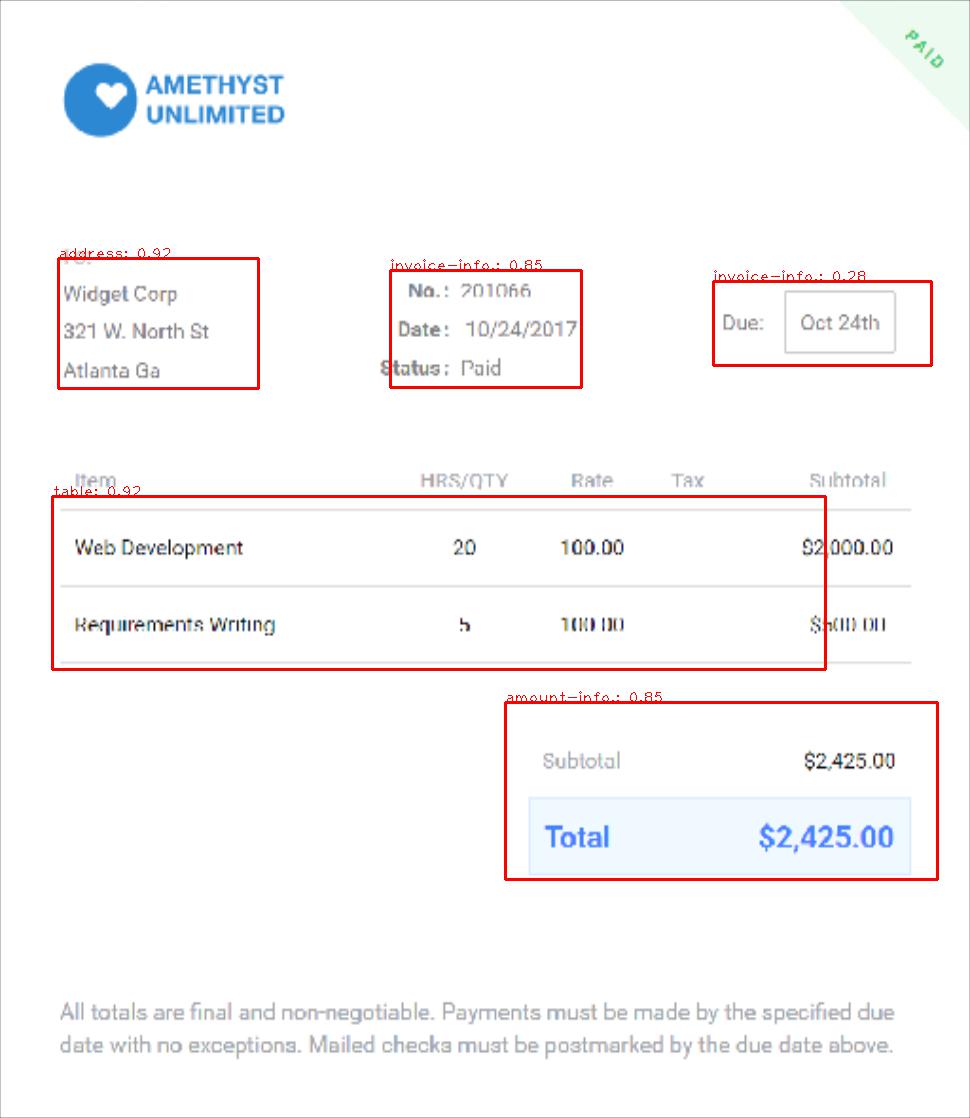}
\includegraphics[width=0.5\textwidth,height=0.5\textheight]{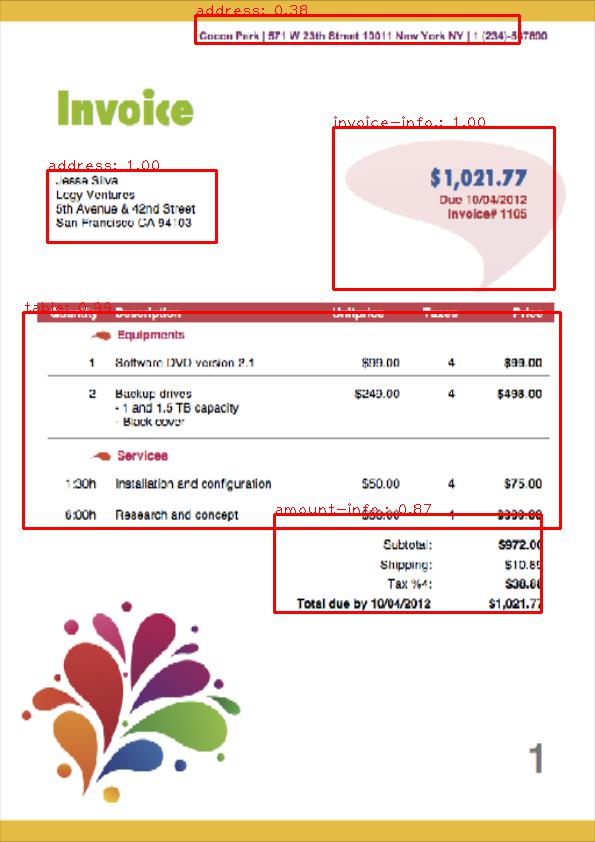}
   \caption{Sample predictions from our system on the test images of Invoice Dataset}
\label{fig:invoices_pred}
\end{figure*}


\section{Conclusion}
In this work, we have shown that object detection techniques can be used for Document Layout understanding. We have also shown that the proposed methodology can be scaled across multiple domains with just need of few tagged examples. The results also demonstrate the superiority of the methodology over existing object detection techniques.

Document Layout analysis techniques assumes great importance in the information age as more and more documents are digitized and needs to be retrieved by understanding their content similar to digital content. Such techniques are useful in automating manually intensive business processes such as processing KYC documents or invoices. Document Layout analysis techniques also opens up the possibilities for businesses to mine documents such as paper receipts and extract valuable insights from them for market research purposes. Getting a large annotated corpus of data can be time-consuming and expensive for practical use-cases which further demonstrates the practical utility of our approach.

{\small
\bibliographystyle{ieee}
\bibliography{egbib}
}

\end{document}